\documentclass[runningheads, orivec]{llncs}
\usepackage[T1]{fontenc}
\usepackage{graphicx}
\usepackage{hyperref}
\usepackage{hhline}
\usepackage{subcaption}
\usepackage{amsmath}
\usepackage{caption}
\usepackage[table,xcdraw]{xcolor}
\usepackage{pdflscape}
\usepackage{rotating}
\usepackage{ifthen}
\usepackage{graphicx}
\usepackage{makecell}
\usepackage{url}
\usepackage{comment}
\usepackage{algorithm2e}

\def\code#1{\texttt{#1}} 
\titlerunning{Node Preservation in CGP}%
\begin{document}

\title{Node Preservation and its Effect on Crossover in Cartesian Genetic Programming}

    \author{Anonymous \and
Anonymous \and
Anonymous}
%

\author{Mark Kocherovsky\orcidID{0000-0002-7313-2325} \and
Illya Bakurov\orcidID{0000-0002-6458-942X} \and Wolfgang Banzhaf\orcidID{0000-0002-6382-3245}}
\authorrunning{M. Kocherovsky et al.}
%
\institute{Department of Computer Science and BEACON Center for the Study of Evolution in Action\\ Michigan State University, East Lansing, MI 48824, USA\\
\email{kocherov@msu.edu}\\
\url{https://banzhaf-lab.github.io/}}

\maketitle

\begin{abstract}
    While crossover is a critical and often indispensable component in other forms of Genetic Programming, such as Linear- and Tree-based, it has consistently been claimed that it deteriorates search performance in CGP.
    As a result, a mutation-alone $(1+\lambda)$ evolutionary strategy has become the canonical approach for CGP. Although several operators have been developed that demonstrate an increased performance over the canonical method, a general solution to the problem is still lacking. In this paper, we compare basic crossover methods, namely one-point and uniform, to variants in which nodes are ``preserved,'' including the subgraph crossover developed by Roman Kalkreuth, the difference being that when ``node preservation'' is active, crossover is not allowed to break apart instructions. We also compare a node mutation operator to the traditional point mutation; the former simply replaces an entire node with a new one. We find that node preservation in both mutation and crossover improves search using symbolic regression benchmark problems, moving the field towards a general solution to CGP crossover.
\end{abstract}

\section{Introduction}
Genetic Programming (GP) has seen several decades of research as a subfield of Evolutionary Computation (EC). Established in the late 1980s / early 1990s by John Koza~\cite{koza1990genetic,koza1994genetic}, GP extends traditional Genetic Algorithmic (GA) methods by evolving \textit{programs}, ranging from basic Boolean and symbolic expressions to complex system control algorithms. 
Programs are traditionally represented as trees (TGP), where leaves contain terminals (input variables and constants) and parent nodes contain operators. As the tree is traversed, an expression is constructed. Crossover occurs by swapping subtrees between individuals, and mutation changes the tree structure. 
A notable feature of GP is the presence of \textit{introns}, also referred to as \textit{ineffective instructions}. These are parts of the program that do not influence the final output behavior, yet they are believed to play a beneficial role by providing additional genetic material for variation operators to act upon. In contrast, \textit{exons} are parts of the code that directly contribute to the output behavior of programs~\cite{brameier2001evolving,nordin1996explicitly}.

A consistent problem with GP is \textit{bloat}, a phenomenon where individuals become increasingly large, thus more complex, leading to slower run-times and deteriorating interpretability. Early studies established that the majority of the genome in GP consists of introns~\cite{c95_complexity_compression_nordin,nordin1996explicitly}. More recent findings have shown that this proportion further increases when also considering genes that only marginally affect behavioral semantics~\cite{banzhaf2024nature}. Additionally, even under exploratory evolutionary dynamics, the genotypic population is often dominated by a small number of distinct phenotypes, i.e., tree structures composed primarily of exonic (active) components.

Bloat in TGP has been extensively studied, with several theories proposed to explain its emergence~\cite{p94_recombination_selection_construction_tackett,c98_fitness_causes_bloat_langdon,c98_fitness_causes_bloat_mutation_langdon,c07_crossover_bias_theory_dignum,c95_accurate_replication_in_gp_mcphee,c94_gp_redundancy_blickle,b94_evolution_of_evolvability_altenberg,a02_causes_of_growth_soule,banzhaf2002some}. One prominent theory derives parallels from intron research in genetics~\cite{c96_introns_genetics_wu}, and suggests that bloat arises as a protective mechanism to preserve useful substructures, or building blocks, which are otherwise vulnerable to disruption by mutation and crossover~\cite{nordin1996explicitly,c95_complexity_compression_nordin,c95_accurate_replication_in_gp_mcphee,c94_gp_redundancy_blickle,b94_evolution_of_evolvability_altenberg,a02_causes_of_growth_soule}. This vulnerability arises from the lack of explicit built-in mechanisms for preservation and modular reuse of building blocks in basic TGP. 

As a response to TGP's limitations, numerous alternative GP paradigms have been developed. These primarily differ in the data structures used to represent candidate solutions and in the design of initialization and variation operators tailored to those representations\cite{banzhaf1998genetic,brameier2007linear}.
One of these alternatives is Linear GP (LGP), where programs are represented as variable-length sequences of instructions executed in order~\cite{banzhaf1998genetic,brameier2007linear,a02_push_gp_spector,a01_grammatical_evolution_oneill,miller1999empirical,miller2008cartesian}.
Each instruction operates on a set of registers, which serve as a steady-state memory module that stores intermediate results. This register-based architecture facilitates the reuse of partial computations and reduces the risk of structural disruption during variation~\cite{kocherovsky2024crossover}. 

Cartesian Genetic Programming (CGP) is another form of GP; CGP uses a fixed-length linear genome to encode programs represented as directed acyclic graphs (DAGs), where each gene specifies a functional component and its connections in a cartesian grid-like structure~\cite{miller2008cartesian}. Each node in the genome applies a function to a fixed number of inputs, selected from earlier nodes or input variables, thereby supporting explicit reuse of intermediate results. Unlike LGP, however, CGP does not use registers and instead relies on positional references within the genome~\cite{kocherovsky2024crossover}. 

The fixed-length representation in CGP helps to control bloat by forcing the evolutionary search to operate within a bounded genomic space. However, this structural rigidity can also limit evolvability, particularly in problems that benefit from larger or more modular solutions. 
On the other hand, the fixed-length encoding enables direct crossover at the genomic level. Thus, it became tempting for researchers and practitioners to use traditional GA-style crossover operators in CGP. Yet, empirical studies have consistently shown that such operators are often detrimental to CGP's search performance~\cite{clegg2007new,kocherovsky2024crossover}, and the scientific community mostly avoids their usage~\cite{a20_cgp_status_future_miller}. As such, a ($1+\lambda$) configuration is typically used in practice. 

In response, researchers have proposed systematic explanations for CGP’s crossover limitations and introduced alternative operators specifically tailored to its representation~\cite{cai2006positional,clegg2007new,c08_advanced_modules_cgp_kaufmann,da2018cartesian,c17_subgraph_xover_cgp_kalkreuth,husa2018comparative,cui2023towards,c23_equidistant_cui,cui2024positioning}. In particular,  Kalkreuth~\cite{c17_subgraph_xover_cgp_kalkreuth,kalkreuth2020comprehensive,kalkreuth2021reconsideration} and Cui~\cite{c23_equidistant_cui,cui2023towards,cui2024positioning,Cui2025Reorder}, have advanced a deeper understanding of CGP's structural and positional constraints, leading to more principled crossover operators. These efforts demonstrate that effective crossover is possible when grounded in a clear rationale, yet the field still lacks a unified theory or widely accepted best practices for recombination in CGP.
In 2024, Kocherovsky and Banzhaf hypothesized that CGP lacks means to ``anchor'' good substructures, as the steady-state memory does in LGP~\cite{kocherovsky2024crossover}.
Building on this, their 2025 work~\cite{kocherovsky2025on} reaffirmed the positional and structural insights from Cui~\cite{cui2023towards}, arguing that the failure of some CGP crossover strategies stems from implicit assumptions about positional equivalence and structural interchangeability. From this basis, we follow Kalkreuth~\cite{kalkreuth2020comprehensive} and Kocherovsky et al.~\cite{kocherovsky2024crossover,kocherovsky2025on} in testing one of the features of Subgraph crossover in the \textit{general} case: crossover in Subgraph crossover \textit{necessarily operates at node-level, instead of at the genomic level}, because otherwise its behavioral constraints would be broken. We call this technique \textbf{Node Preservation}, because the node is structurally preserved throughout the operation.

Here we thus argue that by ensuring to perform node preservation --- which can be accomplished as easily as changing the representation of a program from a one-dimensional vector to a two-dimensional matrix --- both crossover and mutation are \textit{generally} improved. We test one-point and uniform crossover methods, which are  classical operators in the wider field of evolutionary computation both with and without this constraint, each in the typical CGP representation and in a more LGP-like two-dimensional representation, where we find that node preservation significantly improves the likelihood of producing better solutions. We also compare these methods to the Subgraph crossover, which performs better than traditional operators, likely due to the care it takes in making sure subgraphs of different parents manage to link together.



The structure of this paper is as follows. In Section \ref{sec:literature} we discuss the previous literature on CGP crossover. In Section \ref{sec:cgp} we explain the structure of our CGP models and introduce our Node Preservation Method. Section \ref{sec:experiments} explains our experimental conditions, Section \ref{sec:results} introduces and discusses our results, while Section \ref{sec:discussion} concludes the paper with a list of contributions and potential for future work.

We also make our code and supplementary material available on our github here: ANONYMIZED.

\section{Background}\label{sec:literature}

\subsection{GA-like Crossover in CGP}
The use of GA-like fixed-length encoding in CGP facilitates direct crossover at the genomic level, which could justify why CGP researchers and practitioners started to use traditional GA-style recombination methods such as $n$-point and uniform crossover~\cite{oltean2004encoding}. 
In $n$-point crossover, a small number of crossover points results in the exchange of contiguous segments between parents. This preserves the relative ordering of neighboring genes, under the assumption that genes located near one another are functionally related and should be inherited together to maintain their combined effect. As the number of crossover points increases, this assumption weakens. In the extreme case (uniform crossover) each gene is swapped with a probability of 0.5 (typically independently), entirely disregarding positional relationships. Such an approach treats the genome as an unordered set of genes, maximizing diversity at the cost of disrupting structural coherence.

While these assumptions often hold in many GA applications, such as Traveling Salesman Problem~\cite{c20_tsp_gas_sun}, they break down in the context of CGP~\cite{cai2006positional}. CGP encodes programs as directed acyclic graphs (DAGs), where each node can connect to any precedent node in the genome, rather than just to adjacent ones. Consequently, functional relationships between genes are not determined by proximity in the genome, and assuming otherwise can lead to the disruption of essential dependencies during crossover. Moreover, ignoring positional information altogether, as uniform crossover does, can exacerbate the problem by destroying useful structures, particularly since the functionality of a node at index $i$ can vary significantly between parents.
This issue was first raised by Cai \textit{et. al.}~\cite{cai2006positional} and termed as \textit{positional dependence}: \textit{``The effect or meaning of a component in the evolved or resulting program is determined by its absolute or relative position in the program representation''}. 
Later, Goldman and Punch~\cite{goldman2013length} deepened the understanding by introducing the concept of \textit{positional bias} to describe the fact that a node's activation probability varies significantly by position, with a bias toward nodes near the input, which receive more connections due to the DAG structure in CGP, where each node can only connect to preceding nodes to avoid cycles. The presence of positional bias was further confirmed in~\cite{kocherovsky2025on,cui2023towards}. 

To our knowledge, the first attempt to design a better crossover operator for CGP was presented in 2007 by Clegg \textit{et. al.}~\cite{clegg2007new}. The authors, however, essentially adapted the real-valued crossover often used in GA literature for solving continuous optimization problems. This crossover performs a randomly-weighted average of the two parent genes. Given that traditional CGP uses an integer-based representation, the authors proposed an algorithm for converting it into a $[0,1]$ range of floating-points. 
The real-valued CGP crossover was assessed on two symbolic regression problems (Koza2 and Koza3) and was found to outperform the mutation-alone strategy only on one problem. Nevertheless, it was found to achieve faster convergence in earlier iterations, which led the authors to adapt the crossover rate over generations. The adaptive variant turned out to outperform the mutation-alone strategy. However, later studies examined real-valued crossover on more problems and found no advantage in using it. The paper's methodology has also been called into question given a relatively small amount of data presented~\cite{ms12_improving_cgp_turner,kocherovsky2025on}. While Clegg et al. also used indivisible node blocks as crossover units, the authors of Cui et al. (2024) were unable to replicate these results~\cite{cui2024positioning}.
\subsection{Representation-aware Crossover in CGP}
Early efforts using GA-style crossover in CGP often failed to respect the inherent structural dependencies encoded in CGP's DAG. Consequently, a new wave of crossover research has emerged that explicitly accounts for positional dependence and positional bias, while respecting the underlying graph-based representation of CGP.
In 2017, Kalkreuth \textit{et. al.} introduced the Subgraph Crossover, which acts as a one-point crossover where the crossover point is bounded by the range of active nodes in the program~\cite{c17_subgraph_xover_cgp_kalkreuth}. This was found to outperform traditional crossover methods on Boolean and symbolic regression problems~\cite{c17_subgraph_xover_cgp_kalkreuth,kalkreuth2020comprehensive,kalkreuth2021reconsideration}. Husa and Kalkreuth introduced block crossover, which swaps coherent blocks of instructions between parents~\cite{husa2018comparative}. An operator that combined well-performing parent subgraphs into a single chromosome was introduced by da Silva and Bernardino~\cite{da2018cartesian}. 

Goldman and Punch~\cite{goldman2013length} observed that the distribution of active nodes across the CGP genome tends to be biased towards earlier indices and proposed two strategies to mitigate this bias. The first, called \code{DAG}, mutates connections to every other node in the genome, as long as the new connection does not create a cycle; the second, called \code{REORDER}, shuffles active nodes in the genome in such a way that the phenotype does not change. Both strategies led to an increased the amount of active nodes in evolved graphs and achieved faster convergence towards a solution with smaller genotypes.
Following Goldman and Punch, Cui \textit{et. al.} demonstrated that applying the \code{REORDER} operator could mitigate positional bias and enable crossover to contribute positively to CGP search~\cite{cui2023towards}. This work led to further enhancements, such as the Equidistant-\code{REORDER} operator~\cite{c23_equidistant_cui}, which more uniformly distributes exons across the genome. Performing Equidistant-\code{REORDER} was found to improve CGP's problem-solving capabilities across Boolean and regression benchmarks~\cite{cui2024positioning}.

\section{Methods}\label{sec:cgp}
In Cartesian Genetic Programming, a program is defined as a set of nodes, of which there are three types: \textit{input}, including variable and constant terminals; \textit{function}, which take in input and carry out instructions; and \textit{output}, which take in one argument each for the program to return. Typically, a CGP genome is represented as a linear genome in the form of a one-dimensional vector of function and output nodes:
\begin{equation}\label{eq:trad}
    p_i \in P = f_0a_{0,1}a_{0,2}f_1a_{1,1}a_{1,2}...f_na_{n,1}a_{n,2}o_0o_1...o_m
\end{equation}
assuming a program $p\in P$ contains $n$ function nodes, each with one operator $f\in F$ and an arity $a=2$, and $m$ output nodes. In this work, although we follow~\cite{kocherovsky2024crossover} and~\cite{kocherovsky2025on}, we use a slightly different representation: each program is represented as a \textit{matrix}, with each row representing a node. Technically, the matrix would look as shown in Table \ref{tab:program}. Note the recording of which function nodes are active, which is important for crossover algorithms and data analysis; and the separation of input and constant nodes, which is done for clarity.

\vspace{-0.75cm}
\begin{table}[ht]
    \caption{Example CGP program $p$ in this work. The full representation has a length of $n+|I|+|O|$, where there are $n$ function nodes, $|I|$ input and constant nodes, and $|O|$ output nodes. A unified index registry allows for easy reference to and from nodes of different types. }
    \label{tab:program}
    \resizebox{\columnwidth}{!}{%
    \begin{tabular}{|l|l|l|l|l|l|l|}
        \hline
        Index & NodeType & Value & Operator & Operand0 & Operand1 & Active\\ \hline
        0 & InputNode & x & NaN & NaN & NaN & NaN\\ \hline
        1 & ConstantNode & 1 & NaN & NaN & NaN & NaN\\ \hline
        2 & FunctionNode & $x+1$ & Addition & 0 & 1 & 1\\ \hline
        3 & FunctionNode & 0 & Multiplication & 2 & 2 & 0\\ \hline
        ... & ... & ... & ... & ... & ... & ...\\ \hline
        $n+2$ & FunctionNode & 0 & Analytic Quotient & $i$ & $j$ & 0\\ \hline
        $o_0$ & OutputNode & $p(x)$ & NaN & 2 & NaN & NaN\\ \hline
    \end{tabular}%
}
\end{table}
This decision has four motivations: first, it is easier to program and manage on a technical level; second, it is  more readable and writable on the software design level; third, this representation offers the user more information as to the internal dynamics of the model; and finally, it makes it easier to swap nodes as all a practitioner needs to is to swap rows between parents, thus following Kalkreuth~\cite{kalkreuth2021reconsideration}. 

\subsection{Crossover}\label{sec:crossover}

In addition to the canonical $(1+4)$ strategy, we also implement the traditional one-point and uniform crossovers~\cite{oltean2004encoding}. The latter is simply a form of $n$-point crossover, where $n$ is equal to half the length of the genome. While in all replicates, the matrix form in Table \ref{tab:program} is used generally, in runs \textit{without} node preservation, functions are used to convert to and from the more typical form given in Equation \ref{eq:trad} specifically to perform crossover. For all our methods that involve crossover, a (40+40) strategy is implemented, following~\cite{kocherovsky2024crossover} and~\cite{kocherovsky2025on}.

The Subgraph crossover, developed by Kalkreuth~\cite{c17_subgraph_xover_cgp_kalkreuth,kalkreuth2020comprehensive,kalkreuth2021reconsideration}, works in three main steps, assuming an otherwise one-point crossover. First, a crossover point is chosen \textit{within} the range of active nodes. The operator must swap \textit{sub}graphs of the program, hence the range constraint \textit{and} the need to prevent crossover points from breaking apart functions. If the nodes are not preserved, then the subgraph will change during crossover. Second, the parent programs swap \textit{nodes} and are recombined. Finally, the first active node after the crossover point has connections re-wired to ensure that both subgraphs remain active.

\subsection{Mutation}

We test two mutation operators. The first is point mutation, which is standard for CGP. If a model is chosen to be mutated, a single output or function node is chosen at random, and either has its operator or operands changed to a random legal value. The second is \textit{node} mutation, where a randomly chosen node is simply replaced with an entirely new randomly generated node. We test all operators with both mutation types separately to provide a comprehensive comparison.

\subsection{Selection and Fitness}
For selection, we primarily use elite tournament selection, except for canonical ($1+\lambda$) experiments, which necessarily use elitism. Neutral drift is allowed in all cases. Following~\cite{kocherovsky2024crossover} and~\cite{kocherovsky2025on}, our fitness function is based on the Pearson correlation $r$: $f = 1-r^2$~\cite{haut2023correlation}.

\section{Experiments}\label{sec:experiments}

Following~\cite{kocherovsky2024crossover} and~\cite{kocherovsky2025on}, we test our methods on ten symbolic regression problems. These are three Koza problems, four Nguyen problems, the Ackley, Rastrigin, and Levy problems. These are shown in detail in Table \ref{tab:problems}. For the simpler problems, training sets consisted of 20 points and test sets consisted of 10 points; for the complex problems, training sets consisted of 40 points, with test sets of 20 points. 
In all cases, the training-testing split was 0.33. All models were allowed to use the four arithmetic functions, with division handled by Analytic Quotient~\cite{ni2012use}.

\vspace{-0.75cm}
\label{tab:problems}
\begin{table}[ht]
\caption{Problems tested; each model was given a set of random points within the given domain. Mostly reproduced from~\cite{kocherovsky2025on}.}
\centering
    \resizebox{\textwidth}{!}{%
    \begin{tabular}{|l|l|l|l|}
        \Xhline{3\arrayrulewidth}
         \textbf{Problem}& \textbf{Function} & \textbf{Domain} & \textbf{Points}\\
        \hline
        Koza-1 & $x^4+x^3+x^2+x$ & [-1, 1] & 30 \\
        \hline
        Koza-2 & $x^5-2x^3+x$ & [-1, 1] & 30 \\
        \hline
        Koza-3 & $x^6-2x^4+x^2$ & [-1, 1] & 30\\
        \hline
        Nguyen-4 & $x^6+x^5+x^4+x^3+x^2+x$ & [-1, 1] & 30\\
        \hline
        Nguyen-5 & $\sin(x^2)\cos(x)-1$ & [-1, 1] & 30\\
        \hline
        Nguyen-6 & $\sin(x)+\sin(x+x^2)$ & [-1, 1] & 30\\
        \hline
        Nguyen-7 & $\ln(x+1)+\ln(x^2+1)$ & [0, 2] & 30\\
        \hline
        Ackley~\cite{ackley2012connectionist} & $-20\exp(-0.2x^2)-\exp(\cos{2\pi x})+20+\exp{1}$ & [-32.768, 32.768] & 60\\
        \hline
        Rastrigin~\cite{rudolph1990globale} & $10+x^2-10\cos{2\pi x}$ & [-5.12, 5.12] & 60\\
        \hline
        Levy~\cite{laguna2005experimental} & See Reference & [-10, 10] & 60\\
        \hline
    \end{tabular}
    }
\end{table}

Our full list of shorthand used to refer to the various crossover methods is shown in Table \ref{tab:algorithms}. Each method has fifty replicates for each problem, with each replicate run for six thousand generations. Each model is given a maximum size (i.e. number of active nodes) of 64. We collect a series of metrics in each generation: best fitness, median fitness, best model size, median model size, and semantic diversity between each model in the population.

All experiments were run on ANONYMIZED.

\begin{table}[ht]
   \centering
   \caption{Evolutionary parameters to demonstrate the effects of various crossover and mutation strategies. Mostly reproduced from~\cite{kocherovsky2025on}.}
    \begin{tabular}{|l|l|l|}
        \hline
        \textbf{Notation} & \textbf{Crossover} & \textbf{Mutation}\\
        \hline
        CGP(1+4)PM & None (Canonical) & Point (100\%)\\
        \hline
        CGP-1x(40+40)PM & One-Point Nodes Preserved (50\%) & Point (50\%)\\
        \hline
        CGP-1x1d(40+40)PM & One-Point (50\%)  & Point (50\%)\\
        \hline
        CGP-Ux(40+40)PM & Uniform Nodes Preserved (50\%) & Point (50\%) \\
        \hline
        CGP-Ux1d(40+40)PM & Uniform (50\%) & Point (50\%) \\
        \hline
        CGP-SGx(40+40)PM & Subgraph (50\%)  & Point (50\%)\\
        \hline
        CGP(1+4)NM & None (Canonical) & Node (50\%)\\
        \hline
        CGP-1x(40+40)NM & One-Point (50\%) & Node (50\%)\\
        \hline
        CGP-1x1d(40+40)NM & One-Point Flattened (50\%)  & Node (50\%)\\
        \hline
        CGP-Ux(40+40)NM & Uniform (50\%) & Node (50\%)\\
        \hline
        CGP-Ux1d(40+40)NM & Uniform Flattened (50\%) & Node (50\%)\\
        \hline
        CGP-SGx(40+40)NM & Subgraph (50\%)  & Node (50\%)\\
        \hline
    \end{tabular}
    \label{tab:algorithms}
\end{table}

\subsection{Plackett-Luce Analysis}
Following~\cite{c23_equidistant_cui}, we use the Plackett-Luce model to compare our results at the end of evolution and provide a value indicating the probability that a given crossover operator will produce the best model compared to the other operators for a given problem. In this study, we use the \code{PlackettLuce} library in R provided by Turner et al.~\cite{plackettluce}. Because it is based on rankings, it is important to remember that the Plackett-Luce model does not give us the probability that a model will be objectively ``good'', but only the probability that an operator will produce a better model than its competitors. This allows us to make more definite statements when comparing the results of the various operators without having to go back-and-forth between the fitness table and matrices of $p$-values.


\section{Results}\label{sec:results}
\subsection{Fitness}
\begin{table}[ht]
\caption{Median Fitness for the best individual in all 50 replicates. The best operator's fitness in each problem is highlighted in deep green and written in bold, and the worst fitness is highlighted in deep red.}
\label{tab:results}
\resizebox{\textwidth}{!}{%
\begin{tabular}{|l|rr|rr|rr|}
\hline
          & \multicolumn{2}{c|}{\textbf{CGP(1+4)}}                                                  & \multicolumn{2}{c|}{\textbf{CGP-1X(40+40)}}                                             & \multicolumn{2}{c|}{\textbf{CGP-1x1d(40+40)}}                                           \\
          & \multicolumn{1}{c}{Point Mutation}         & \multicolumn{1}{c|}{Node Mutation}         & \multicolumn{1}{c}{Point Mutation}         & \multicolumn{1}{c|}{Node Mutation}         & \multicolumn{1}{c}{Point Mutation}         & \multicolumn{1}{c|}{Node Mutation}         \\ \hline
Koza 1    & \cellcolor[HTML]{E67C73}1.323E-02          & \cellcolor[HTML]{FAC468}3.288E-03          & \cellcolor[HTML]{57BB8A}\textbf{3.389E-04} & \cellcolor[HTML]{98C57C}5.052E-04          & \cellcolor[HTML]{FFD666}8.183E-04          & \cellcolor[HTML]{FFD566}9.353E-04          \\
Koza 2    & \cellcolor[HTML]{E88172}2.368E-02          & \cellcolor[HTML]{E67C73}2.444E-02          & \cellcolor[HTML]{C2CC73}5.897E-03          & \cellcolor[HTML]{73BF84}3.126E-03          & \cellcolor[HTML]{F2A46D}1.720E-02          & \cellcolor[HTML]{F5B16B}1.480E-02          \\
Koza 3    & \cellcolor[HTML]{E67C73}7.156E-02          & \cellcolor[HTML]{F9BE69}3.323E-02          & \cellcolor[HTML]{CCCD71}1.474E-02          & \cellcolor[HTML]{A5C77A}1.166E-02          & \cellcolor[HTML]{FAC468}2.952E-02          & \cellcolor[HTML]{FED266}2.155E-02          \\
Nguyen 4  & \cellcolor[HTML]{E67C73}4.207E-03          & \cellcolor[HTML]{EA8871}3.803E-03          & \cellcolor[HTML]{F2D369}8.881E-04          & \cellcolor[HTML]{5CBB89}5.530E-04          & \cellcolor[HTML]{F9BF69}1.782E-03          & \cellcolor[HTML]{E1D16D}8.516E-04          \\
Nguyen 5  & \cellcolor[HTML]{E67C73}1.469E-03          & \cellcolor[HTML]{F8BC69}5.610E-04          & \cellcolor[HTML]{FFD666}1.849E-04          & \cellcolor[HTML]{57BB8A}\textbf{3.732E-05} & \cellcolor[HTML]{F6B46A}6.693E-04          & \cellcolor[HTML]{FCC967}3.764E-04          \\
Nguyen 6  & \cellcolor[HTML]{E67C73}2.376E-03          & \cellcolor[HTML]{F7B86A}1.106E-03          & \cellcolor[HTML]{57BB8A}\textbf{2.297E-04} & \cellcolor[HTML]{77C083}2.706E-04          & \cellcolor[HTML]{FFD466}4.852E-04          & \cellcolor[HTML]{FDCF67}6.026E-04          \\
Nguyen 7  & \cellcolor[HTML]{F8BA6A}1.345E-04          & \cellcolor[HTML]{FFD566}1.920E-05          & \cellcolor[HTML]{7DC182}5.111E-06          & \cellcolor[HTML]{A3C77A}7.018E-06          & \cellcolor[HTML]{E67C73}4.012E-04          & \cellcolor[HTML]{FFD466}2.060E-05          \\
Ackley    & \cellcolor[HTML]{E98671}4.920E-02          & \cellcolor[HTML]{E67C73}5.113E-02          & \cellcolor[HTML]{FCCB67}3.520E-02          & \cellcolor[HTML]{FAC368}3.678E-02          & \cellcolor[HTML]{FFD366}3.349E-02          & \cellcolor[HTML]{B2C977}3.161E-02          \\
Levy      & \cellcolor[HTML]{E67C73}1.383E-01          & \cellcolor[HTML]{F09F6E}1.225E-01          & \cellcolor[HTML]{84C281}8.726E-02          & \cellcolor[HTML]{C9CD72}9.271E-02          & \cellcolor[HTML]{FBD567}9.657E-02          & \cellcolor[HTML]{F5D469}9.611E-02          \\
Rastrigin & \cellcolor[HTML]{F1A16D}5.315E-01          & \cellcolor[HTML]{E67C73}5.648E-01          & \cellcolor[HTML]{CCCD71}4.688E-01          & \cellcolor[HTML]{F8D468}4.801E-01          & \cellcolor[HTML]{FDCE67}4.900E-01          & \cellcolor[HTML]{E7D26B}4.758E-01          \\ \hline
          & \multicolumn{2}{c|}{\textbf{CGP-Ux(40+40)}}                                             & \multicolumn{2}{c|}{\textbf{CGP-Ux1d(40+40)}}                                           & \multicolumn{2}{c|}{\textbf{CGP-SGx(40+40)}}                                            \\
          & \multicolumn{1}{c}{Point Mutation}         & \multicolumn{1}{c|}{Node Mutation}         & \multicolumn{1}{c}{Point Mutation}         & \multicolumn{1}{c|}{Node Mutation}         & \multicolumn{1}{c}{Point Mutation}         & \multicolumn{1}{c|}{Node Mutation}         \\ \hline
Koza 1    & \cellcolor[HTML]{92C47E}4.894E-04          & \cellcolor[HTML]{E9D26B}7.112E-04          & \cellcolor[HTML]{FED166}1.562E-03          & \cellcolor[HTML]{FFD466}1.075E-03          & \cellcolor[HTML]{5BBB89}3.512E-04          & \cellcolor[HTML]{5FBC89}3.614E-04          \\
Koza 2    & \cellcolor[HTML]{ACC878}5.107E-03          & \cellcolor[HTML]{C5CC73}5.990E-03          & \cellcolor[HTML]{FCCB67}1.005E-02          & \cellcolor[HTML]{FCCB67}1.001E-02          & \cellcolor[HTML]{57BB8A}\textbf{2.139E-03} & \cellcolor[HTML]{6BBE86}2.841E-03          \\
Koza 3    & \cellcolor[HTML]{DBD06E}1.597E-02          & \cellcolor[HTML]{57BB8A}\textbf{5.382E-03} & \cellcolor[HTML]{ED946F}5.750E-02          & \cellcolor[HTML]{FDCD67}2.445E-02          & \cellcolor[HTML]{77C084}7.979E-03          & \cellcolor[HTML]{9FC67B}1.115E-02          \\
Nguyen 4  & \cellcolor[HTML]{57BB8A}\textbf{5.400E-04} & \cellcolor[HTML]{6EBE86}5.920E-04          & \cellcolor[HTML]{FBC868}1.444E-03          & \cellcolor[HTML]{FFD666}9.458E-04          & \cellcolor[HTML]{FDCC67}1.294E-03          & \cellcolor[HTML]{AEC978}7.360E-04          \\
Nguyen 5  & \cellcolor[HTML]{AFC977}1.133E-04          & \cellcolor[HTML]{94C47D}9.046E-05          & \cellcolor[HTML]{FAC468}4.491E-04          & \cellcolor[HTML]{FAD567}1.778E-04          & \cellcolor[HTML]{68BD87}5.235E-05          & \cellcolor[HTML]{6DBE86}5.637E-05          \\
Nguyen 6  & \cellcolor[HTML]{67BD87}2.499E-04          & \cellcolor[HTML]{D9D06E}3.928E-04          & \cellcolor[HTML]{FECF67}5.926E-04          & \cellcolor[HTML]{FBC668}7.864E-04          & \cellcolor[HTML]{95C47D}3.071E-04          & \cellcolor[HTML]{87C280}2.903E-04          \\
Nguyen 7  & \cellcolor[HTML]{C4CC73}8.678E-06          & \cellcolor[HTML]{A6C77A}7.140E-06          & \cellcolor[HTML]{FFD566}1.820E-05          & \cellcolor[HTML]{FFD666}1.450E-05          & \cellcolor[HTML]{57BB8A}\textbf{3.169E-06} & \cellcolor[HTML]{85C281}5.500E-06          \\
Ackley    & \cellcolor[HTML]{FDCE67}3.466E-02          & \cellcolor[HTML]{7DC182}3.076E-02          & \cellcolor[HTML]{C8CD72}3.197E-02          & \cellcolor[HTML]{D6CF6F}3.220E-02          & \cellcolor[HTML]{CBCD71}3.202E-02          & \cellcolor[HTML]{57BB8A}\textbf{3.015E-02} \\
Levy      & \cellcolor[HTML]{57BB8A}\textbf{8.370E-02} & \cellcolor[HTML]{FFD366}9.839E-02          & \cellcolor[HTML]{FFD366}9.829E-02          & \cellcolor[HTML]{FFD666}9.720E-02          & \cellcolor[HTML]{C0CB74}9.196E-02          & \cellcolor[HTML]{FDCF67}1.002E-01          \\
Rastrigin & \cellcolor[HTML]{88C380}4.512E-01          & \cellcolor[HTML]{FDCC67}4.911E-01          & \cellcolor[HTML]{FFD566}4.834E-01          & \cellcolor[HTML]{FBC868}4.953E-01          & \cellcolor[HTML]{BDCB74}4.649E-01          & \cellcolor[HTML]{57BB8A}\textbf{4.383E-01} \\ \hline
\end{tabular}
}
\end{table}

\begin{figure}[ht]
    \centering
    \includegraphics[width=\textwidth]{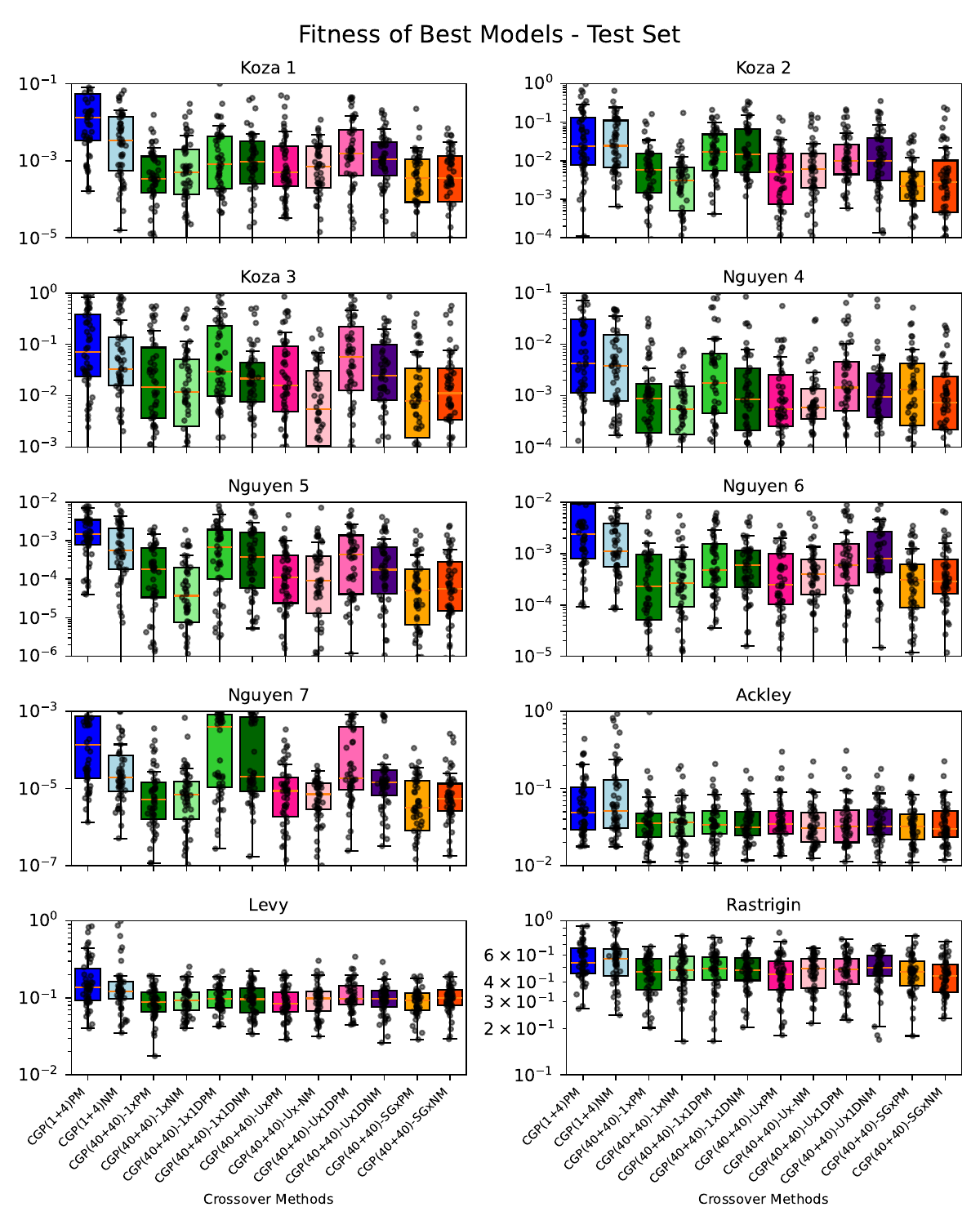}
    \caption{Box plots of best fitness at the end of evolution for each run.}
    \label{fig:fitness_box}
\end{figure}

The median value of the best fitness at the end of each run per crossover operator and problem is shown in Table \ref{tab:results}. Additionally, box plots showing the best fitness at the end of each run are shown in Figure \ref{fig:fitness_box}.

It is clear from the data that Canonical CGP performs significantly worse than other methods, even traditional one-point and uniform methods when the opposite is expected from the reports in most of the literature. This has also been reported in~\cite{cui2024positioning}. Without forming any definite conjectures, we regard this as a point of future investigation. It is also visible that by preserving nodes in crossover and in mutation, we can achieve better results than if instructions were allowed to be broken apart -- or, more accurately, \textit{partially altered} by either operator.

Table \ref{tab:plackett_luce} shows the $p_{\text{best}}$ values for each operator both for each problem and averaged over all problems. This latter table is incredibly useful, as we can see that on average, Subgraph Crossover (which preserves nodes by default) is the most likely to produce the best model across all operators. Some basic statistical operations, if $\langle p(x)\rangle$ is the average probability that operator $x$ produces the best model, demonstrate the following:
\begin{enumerate}
    \item $\langle p(\text{Any SGX})\rangle = 0.246$
    \item $\langle p(\text{Any Full Node Mutation})\rangle = 0.520$
    \item conversely, $\langle p(\text{Any Point Mutation})\rangle = 0.480$
\end{enumerate}
It is therefore more likely that by fully replacing nodes during mutation we can produce better models than by using standard point mutation. Given that on average, the only case where this does not hold is Subgraph Crossover (and even in this case is very close to a 49\%-51\% split), we conjecture that our proposed mutation operator approaches a general solution. It is also clear that node preservation in crossover also seems to present a near-general solution. There are a couple of specific cases where node preservation is \textit{less} likely to produce a better result, but the differences in these rare cases are very small.
\begin{table}[ht]
\centering
\caption{Probability that a crossover operator would result in the best model across all operators both per-problem and averaged over all problems.}
\label{tab:plackett_luce}
\resizebox{\textwidth}{!}{%
\begin{tabular}{|l|rr|rr|rr|}
\hline
          & \multicolumn{2}{c|}{\textbf{CGP(1+4)}}                                      & \multicolumn{2}{c|}{\textbf{CGP-1X(40+40)}}                                 & \multicolumn{2}{c|}{\textbf{CGP-1x1d(40+40)}}                               \\
          & \multicolumn{1}{c}{Point Mutation} & \multicolumn{1}{c|}{Node Mutation}     & \multicolumn{1}{c}{Point Mutation} & \multicolumn{1}{c|}{Node Mutation}     & \multicolumn{1}{c}{Point Mutation}     & \multicolumn{1}{c|}{Node Mutation} \\ \hline
Koza 1    & \cellcolor[HTML]{FFFFFF}0.020      & \cellcolor[HTML]{E8F6EF}0.039          & \cellcolor[HTML]{8BD0AF}0.115      & \cellcolor[HTML]{A2DABF}0.097          & \cellcolor[HTML]{CBEADB}0.063          & \cellcolor[HTML]{C0E6D3}0.073      \\
Koza 2    & \cellcolor[HTML]{FFFFFF}0.035      & \cellcolor[HTML]{F9FDFB}0.040          & \cellcolor[HTML]{BBE4CF}0.086      & \cellcolor[HTML]{6CC499}0.144          & \cellcolor[HTML]{EBF7F1}0.050          & \cellcolor[HTML]{F7FCF9}0.042      \\
Koza 3    & \cellcolor[HTML]{FFFFFF}0.042      & \cellcolor[HTML]{F0F9F5}0.052          & \cellcolor[HTML]{C9E9D9}0.076      & \cellcolor[HTML]{A1D9BE}0.101          & \cellcolor[HTML]{E8F6EF}0.057          & \cellcolor[HTML]{BFE5D2}0.082      \\
Nguyen 4  & \cellcolor[HTML]{FFFFFF}0.037      & \cellcolor[HTML]{F4FBF8}0.043          & \cellcolor[HTML]{9ED8BC}0.096      & \cellcolor[HTML]{57BB8A}\textbf{0.140} & \cellcolor[HTML]{E0F3EA}0.056          & \cellcolor[HTML]{C0E6D3}0.075      \\
Nguyen 5  & \cellcolor[HTML]{FFFFFF}0.023      & \cellcolor[HTML]{E9F6F0}0.041          & \cellcolor[HTML]{AFDFC8}0.088      & \cellcolor[HTML]{57BB8A}\textbf{0.159} & \cellcolor[HTML]{DFF2E9}0.049          & \cellcolor[HTML]{E4F4ED}0.045      \\
Nguyen 6  & \cellcolor[HTML]{FFFFFF}0.032      & \cellcolor[HTML]{F1FAF6}0.039          & \cellcolor[HTML]{6EC49A}0.107      & \cellcolor[HTML]{57BB8A}\textbf{0.118} & \cellcolor[HTML]{ABDDC4}0.075          & \cellcolor[HTML]{A2D9BE}0.080      \\
Nguyen 7  & \cellcolor[HTML]{FFFFFF}0.026      & \cellcolor[HTML]{E0F3EA}0.050          & \cellcolor[HTML]{81CCA7}0.124      & \cellcolor[HTML]{91D3B2}0.112          & \cellcolor[HTML]{FFFFFF}0.026          & \cellcolor[HTML]{F5FBF8}0.034      \\
Ackley    & \cellcolor[HTML]{FAFDFC}0.028      & \cellcolor[HTML]{FFFFFF}0.025          & \cellcolor[HTML]{86CEAB}0.090      & \cellcolor[HTML]{96D5B6}0.081          & \cellcolor[HTML]{8FD2B1}0.085          & \cellcolor[HTML]{7ECBA5}0.094      \\
Levy      & \cellcolor[HTML]{FFFFFF}0.030      & \cellcolor[HTML]{DBF1E6}0.048          & \cellcolor[HTML]{60BF91}0.108      & \cellcolor[HTML]{7CCAA4}0.095          & \cellcolor[HTML]{8ED2B1}0.086          & \cellcolor[HTML]{AADDC4}0.072      \\
Rastrigin & \cellcolor[HTML]{FFFFFF}0.033      & \cellcolor[HTML]{FBFEFC}0.036          & \cellcolor[HTML]{84CEAA}0.111      & \cellcolor[HTML]{CEEBDD}0.064          & \cellcolor[HTML]{B5E1CC}0.080          & \cellcolor[HTML]{A6DBC1}0.089      \\
\hline
Average   & \cellcolor[HTML]{FFFFFF}0.030      & \cellcolor[HTML]{EDF8F2}0.041          & \cellcolor[HTML]{86CEAB}0.100      & \cellcolor[HTML]{73C79E}0.111          & \cellcolor[HTML]{C7E9D8}0.063          & \cellcolor[HTML]{BDE5D1}0.069      \\
Std. Dev. & \cellcolor[HTML]{FFFFFF}0.007      & \cellcolor[HTML]{FFFFFF}0.008          & \cellcolor[HTML]{FFFFFF}0.015      & \cellcolor[HTML]{FFFFFF}0.030          & \cellcolor[HTML]{FFFFFF}0.019          & \cellcolor[HTML]{FFFFFF}0.021      \\ \hline
          & \multicolumn{2}{c|}{\textbf{CGP-Ux(40+40)}}                                 & \multicolumn{2}{c|}{\textbf{CGP-Ux1d(40+40)}}                               & \multicolumn{2}{c|}{\textbf{CGP-SGx(40+40)}}                                \\
          & \multicolumn{1}{c}{Point Mutation} & \multicolumn{1}{c|}{Node Mutation}     & \multicolumn{1}{c}{Point Mutation} & \multicolumn{1}{c|}{Node Mutation}     & \multicolumn{1}{c}{Point Mutation}     & \multicolumn{1}{c|}{Node Mutation} \\ \hline
Koza 1    & \cellcolor[HTML]{ADDEC6}0.088      & \cellcolor[HTML]{A4DAC0}0.095          & \cellcolor[HTML]{D0ECDF}0.059      & \cellcolor[HTML]{C0E6D3}0.072          & \cellcolor[HTML]{57BB8A}\textbf{0.158} & \cellcolor[HTML]{84CEAA}0.121      \\
Koza 2    & \cellcolor[HTML]{A4DABF}0.103      & \cellcolor[HTML]{C3E7D6}0.080          & \cellcolor[HTML]{D6EFE2}0.066      & \cellcolor[HTML]{DEF2E8}0.060          & \cellcolor[HTML]{57BB8A}\textbf{0.159} & \cellcolor[HTML]{77C8A1}0.136      \\
Koza 3    & \cellcolor[HTML]{C3E7D5}0.080      & \cellcolor[HTML]{57BB8A}\textbf{0.147} & \cellcolor[HTML]{F7FCF9}0.047      & \cellcolor[HTML]{C7E9D8}0.077          & \cellcolor[HTML]{87CFAC}0.117          & \cellcolor[HTML]{81CCA8}0.121      \\
Nguyen 4  & \cellcolor[HTML]{9ED8BB}0.097      & \cellcolor[HTML]{7ECBA6}0.116          & \cellcolor[HTML]{D0ECDF}0.066      & \cellcolor[HTML]{B3E1CB}0.083          & \cellcolor[HTML]{B2E0C9}0.084          & \cellcolor[HTML]{8CD1AF}0.107      \\
Nguyen 5  & \cellcolor[HTML]{ADDEC6}0.089      & \cellcolor[HTML]{96D5B6}0.108          & \cellcolor[HTML]{E2F4EB}0.046      & \cellcolor[HTML]{BAE3CF}0.079          & \cellcolor[HTML]{69C397}0.144          & \cellcolor[HTML]{7AC9A2}0.131      \\
Nguyen 6  & \cellcolor[HTML]{70C59C}0.105      & \cellcolor[HTML]{70C59B}0.106          & \cellcolor[HTML]{C1E6D4}0.064      & \cellcolor[HTML]{D1EDDF}0.056          & \cellcolor[HTML]{5EBE8F}0.115          & \cellcolor[HTML]{71C69C}0.105      \\
Nguyen 7  & \cellcolor[HTML]{9DD8BB}0.103      & \cellcolor[HTML]{69C397}0.143          & \cellcolor[HTML]{E8F6EF}0.044      & \cellcolor[HTML]{CBEADB}0.066          & \cellcolor[HTML]{57BB8A}\textbf{0.157} & \cellcolor[HTML]{8BD0AE}0.116      \\
Ackley    & \cellcolor[HTML]{A5DBC0}0.074      & \cellcolor[HTML]{70C59B}0.102          & \cellcolor[HTML]{72C69D}0.100      & \cellcolor[HTML]{80CCA7}0.093          & \cellcolor[HTML]{57BB8A}\textbf{0.115} & \cellcolor[HTML]{5DBE8F}0.111      \\
Levy      & \cellcolor[HTML]{62C092}0.108      & \cellcolor[HTML]{8CD1AF}0.087          & \cellcolor[HTML]{AADDC4}0.072      & \cellcolor[HTML]{96D5B6}0.082          & \cellcolor[HTML]{57BB8A}\textbf{0.113} & \cellcolor[HTML]{6EC59A}0.101      \\
Rastrigin & \cellcolor[HTML]{8AD0AE}0.107      & \cellcolor[HTML]{B4E1CB}0.081          & \cellcolor[HTML]{B0DFC8}0.083      & \cellcolor[HTML]{C2E6D4}0.072          & \cellcolor[HTML]{8CD1AF}0.106          & \cellcolor[HTML]{57BB8A}0.139      \\
\hline
Average   & \cellcolor[HTML]{8ED2B1}0.095      & \cellcolor[HTML]{7BCAA3}0.106          & \cellcolor[HTML]{C4E7D6}0.065      & \cellcolor[HTML]{B3E1CA}0.074          & \cellcolor[HTML]{57BB8A}\textbf{0.127} & \cellcolor[HTML]{65C194}0.119      \\
Std. Dev. & \cellcolor[HTML]{FFFFFF}0.012      & \cellcolor[HTML]{FFFFFF}0.024          & \cellcolor[HTML]{FFFFFF}0.017      & \cellcolor[HTML]{FFFFFF}0.011          & \cellcolor[HTML]{FFFFFF}0.026          & \cellcolor[HTML]{FFFFFF}0.013      \\ \hline
\end{tabular}
}
\end{table}
\vspace{-0.2cm}

\subsection{Complexity}
In Table \ref{tab:size_probabilities}, we show that in most cases the Canonical CGP method is more likely to produce models with fewer active nodes, with one-dimensional crossover methods in distant second. However, this should not be conflated with the probability of finding an objectively better or less complex model, but only in comparison to the other tested methods.
\begin{table}[ht]
\caption{Probability that a given crossover operator will produce the smallest model across all operators.}
\label{tab:size_probabilities}
\resizebox{\textwidth}{!}{%
\begin{tabular}{|l|rr|rr|rr|}
\hline
          & \multicolumn{2}{c|}{\textbf{CGP(1+4)}}                                          & \multicolumn{2}{c|}{\textbf{CGP-1X(40+40)}}                                     & \multicolumn{2}{c|}{\textbf{CGP-1x1d(40+40)}}                               \\
          & \multicolumn{1}{c}{Point Mutation}     & \multicolumn{1}{c|}{Node Mutation}     & \multicolumn{1}{c}{Point Mutation}     & \multicolumn{1}{c|}{Node Mutation}     & \multicolumn{1}{c}{Point Mutation} & \multicolumn{1}{c|}{Node Mutation}     \\ \hline
Koza 1    & \cellcolor[HTML]{57BB8A}\textbf{0.150} & \cellcolor[HTML]{A4DBC0}0.096          & \cellcolor[HTML]{FFFFFF}0.031          & \cellcolor[HTML]{D0ECDF}0.065          & \cellcolor[HTML]{C9EADA}0.069      & \cellcolor[HTML]{94D4B5}0.107          \\
Koza 2    & \cellcolor[HTML]{90D2B2}0.105          & \cellcolor[HTML]{9CD7BB}0.096          & \cellcolor[HTML]{FFFFFF}0.025          & \cellcolor[HTML]{D1EDDF}0.058          & \cellcolor[HTML]{69C397}0.132      & \cellcolor[HTML]{8ED2B0}0.106          \\
Koza 3    & \cellcolor[HTML]{A1D9BE}0.080          & \cellcolor[HTML]{5DBE8E}0.111          & \cellcolor[HTML]{FFFFFF}0.037          & \cellcolor[HTML]{A6DBC1}0.077          & \cellcolor[HTML]{8CD1AF}0.090      & \cellcolor[HTML]{57BB8A}\textbf{0.114} \\
Nguyen 4  & \cellcolor[HTML]{57BB8A}\textbf{0.143} & \cellcolor[HTML]{9CD7BA}0.099          & \cellcolor[HTML]{FFFFFF}0.036          & \cellcolor[HTML]{E5F5ED}0.053          & \cellcolor[HTML]{98D6B7}0.102      & \cellcolor[HTML]{B1E0C9}0.086          \\
Nguyen 5  & \cellcolor[HTML]{57BB8A}\textbf{0.133} & \cellcolor[HTML]{71C69C}0.119          & \cellcolor[HTML]{FBFEFC}0.040          & \cellcolor[HTML]{EFF9F4}0.047          & \cellcolor[HTML]{6FC59B}0.120      & \cellcolor[HTML]{9CD7BA}0.094          \\
Nguyen 6  & \cellcolor[HTML]{79C9A2}0.124          & \cellcolor[HTML]{57BB8A}\textbf{0.146} & \cellcolor[HTML]{FFFFFF}0.034          & \cellcolor[HTML]{CFECDE}0.067          & \cellcolor[HTML]{B4E1CB}0.085      & \cellcolor[HTML]{AFDFC8}0.088          \\
Nguyen 7  & \cellcolor[HTML]{B3E1CA}0.100          & \cellcolor[HTML]{B6E2CC}0.098          & \cellcolor[HTML]{FFFFFF}0.033          & \cellcolor[HTML]{EEF9F3}0.048          & \cellcolor[HTML]{AFDFC8}0.103      & \cellcolor[HTML]{57BB8A}\textbf{0.180} \\
Ackley    & \cellcolor[HTML]{57BB8A}0.119          & \cellcolor[HTML]{58BC8B}0.119          & \cellcolor[HTML]{FCFEFD}0.042          & \cellcolor[HTML]{F1FAF6}0.047          & \cellcolor[HTML]{B2E0C9}0.077      & \cellcolor[HTML]{58BC8B}0.119          \\
Levy      & \cellcolor[HTML]{7CCAA4}0.148          & \cellcolor[HTML]{57BB8A}\textbf{0.180} & \cellcolor[HTML]{FFFFFF}0.031          & \cellcolor[HTML]{F9FDFB}0.037          & \cellcolor[HTML]{B3E0CA}0.099      & \cellcolor[HTML]{ABDDC5}0.106          \\
Rastrigin & \cellcolor[HTML]{57BB8A}0.197          & \cellcolor[HTML]{7BCAA3}0.160          & \cellcolor[HTML]{FDFFFE}0.025          & \cellcolor[HTML]{DDF2E7}0.058          & \cellcolor[HTML]{BEE5D2}0.090      & \cellcolor[HTML]{C2E6D4}0.087          \\ \hline
Average   & \cellcolor[HTML]{57BB8A}\textbf{0.130} & \cellcolor[HTML]{65C194}0.122          & \cellcolor[HTML]{FFFFFF}0.033          & \cellcolor[HTML]{D9F0E4}0.056          & \cellcolor[HTML]{91D3B3}0.097      & \cellcolor[HTML]{7DCBA4}0.109          \\
Std. Dev. & 0.033                                  & 0.030                                  & 0.006                                  & 0.012                                  & 0.019                              & 0.028                                  \\ \hline
          & \multicolumn{2}{c|}{\textbf{CGP-Ux(40+40)}}                                     & \multicolumn{2}{c|}{\textbf{CGP-Ux1d(40+40)}}                                   & \multicolumn{2}{c|}{\textbf{CGP-SGx(40+40)}}                                \\
          & \multicolumn{1}{c}{Point Mutation}     & \multicolumn{1}{c|}{Node Mutation}     & \multicolumn{1}{c}{Point Mutation}     & \multicolumn{1}{c|}{Node Mutation}     & \multicolumn{1}{c}{Point Mutation} & \multicolumn{1}{c|}{Node Mutation}     \\ \hline
Koza 1    & \cellcolor[HTML]{F4FBF8}0.039          & \cellcolor[HTML]{D6EFE3}0.060          & \cellcolor[HTML]{8BD1AF}0.113          & \cellcolor[HTML]{90D2B2}0.110          & \cellcolor[HTML]{BEE5D2}0.077      & \cellcolor[HTML]{B7E2CD}0.082          \\
Koza 2    & \cellcolor[HTML]{F6FCF9}0.032          & \cellcolor[HTML]{DAF0E6}0.052          & \cellcolor[HTML]{C3E7D5}0.068          & \cellcolor[HTML]{57BB8A}\textbf{0.145} & \cellcolor[HTML]{A9DDC3}0.086      & \cellcolor[HTML]{9CD7BA}0.096          \\
Koza 3    & \cellcolor[HTML]{D9F0E5}0.054          & \cellcolor[HTML]{B4E1CB}0.071          & \cellcolor[HTML]{BDE5D1}0.067          & \cellcolor[HTML]{67C295}0.106          & \cellcolor[HTML]{9FD9BC}0.081      & \cellcolor[HTML]{5ABD8C}0.112          \\
Nguyen 4  & \cellcolor[HTML]{FBFEFC}0.039          & \cellcolor[HTML]{D1EDDF}0.065          & \cellcolor[HTML]{84CEAA}0.114          & \cellcolor[HTML]{ACDEC5}0.089          & \cellcolor[HTML]{E9F6F0}0.050      & \cellcolor[HTML]{74C79E}0.125          \\
Nguyen 5  & \cellcolor[HTML]{FFFFFF}0.038          & \cellcolor[HTML]{F0F9F5}0.046          & \cellcolor[HTML]{96D5B6}0.098          & \cellcolor[HTML]{ACDEC6}0.085          & \cellcolor[HTML]{BAE3CF}0.078      & \cellcolor[HTML]{8FD2B1}0.102          \\
Nguyen 6  & \cellcolor[HTML]{F0F9F5}0.045          & \cellcolor[HTML]{DAF0E6}0.059          & \cellcolor[HTML]{B9E3CE}0.082          & \cellcolor[HTML]{98D6B7}0.103          & \cellcolor[HTML]{D7EFE3}0.062      & \cellcolor[HTML]{95D4B5}0.105          \\
Nguyen 7  & \cellcolor[HTML]{F6FCF9}0.041          & \cellcolor[HTML]{E9F7F0}0.053          & \cellcolor[HTML]{ACDEC6}0.106          & \cellcolor[HTML]{A0D9BD}0.116          & \cellcolor[HTML]{F0F9F4}0.047      & \cellcolor[HTML]{D1EDDF}0.074          \\
Ackley    & \cellcolor[HTML]{FFFFFF}0.040          & \cellcolor[HTML]{E3F4EC}0.053          & \cellcolor[HTML]{5CBD8E}\textbf{0.117} & \cellcolor[HTML]{60BF91}0.115          & \cellcolor[HTML]{D7EFE3}0.059      & \cellcolor[HTML]{8FD2B1}0.093          \\
Levy      & \cellcolor[HTML]{FCFEFD}0.034          & \cellcolor[HTML]{D8EFE4}0.066          & \cellcolor[HTML]{B8E3CE}0.094          & \cellcolor[HTML]{BAE3CF}0.093          & \cellcolor[HTML]{E6F5EE}0.053      & \cellcolor[HTML]{DEF2E8}0.060          \\
Rastrigin & \cellcolor[HTML]{FFFFFF}0.022          & \cellcolor[HTML]{E9F7F0}0.045          & \cellcolor[HTML]{C6E8D8}0.082          & \cellcolor[HTML]{AADDC4}0.111          & \cellcolor[HTML]{EEF8F3}0.041      & \cellcolor[HTML]{C8E9D9}0.080          \\ \hline
Average   & \cellcolor[HTML]{F7FCF9}0.038          & \cellcolor[HTML]{D6EFE3}0.057          & \cellcolor[HTML]{96D5B6}0.094          & \cellcolor[HTML]{7FCBA6}0.107          & \cellcolor[HTML]{CBEADB}0.063      & \cellcolor[HTML]{98D6B7}0.093          \\
Std. Dev. & 0.008                                  & 0.009                                  & 0.019                                  & 0.017                                  & 0.016                              & 0.019                                  \\ \hline
\end{tabular}
}
\end{table}

\vspace{-0.5cm}

\begin{figure}[ht]
    \centering
    \centering
    \includegraphics[width=\textwidth]{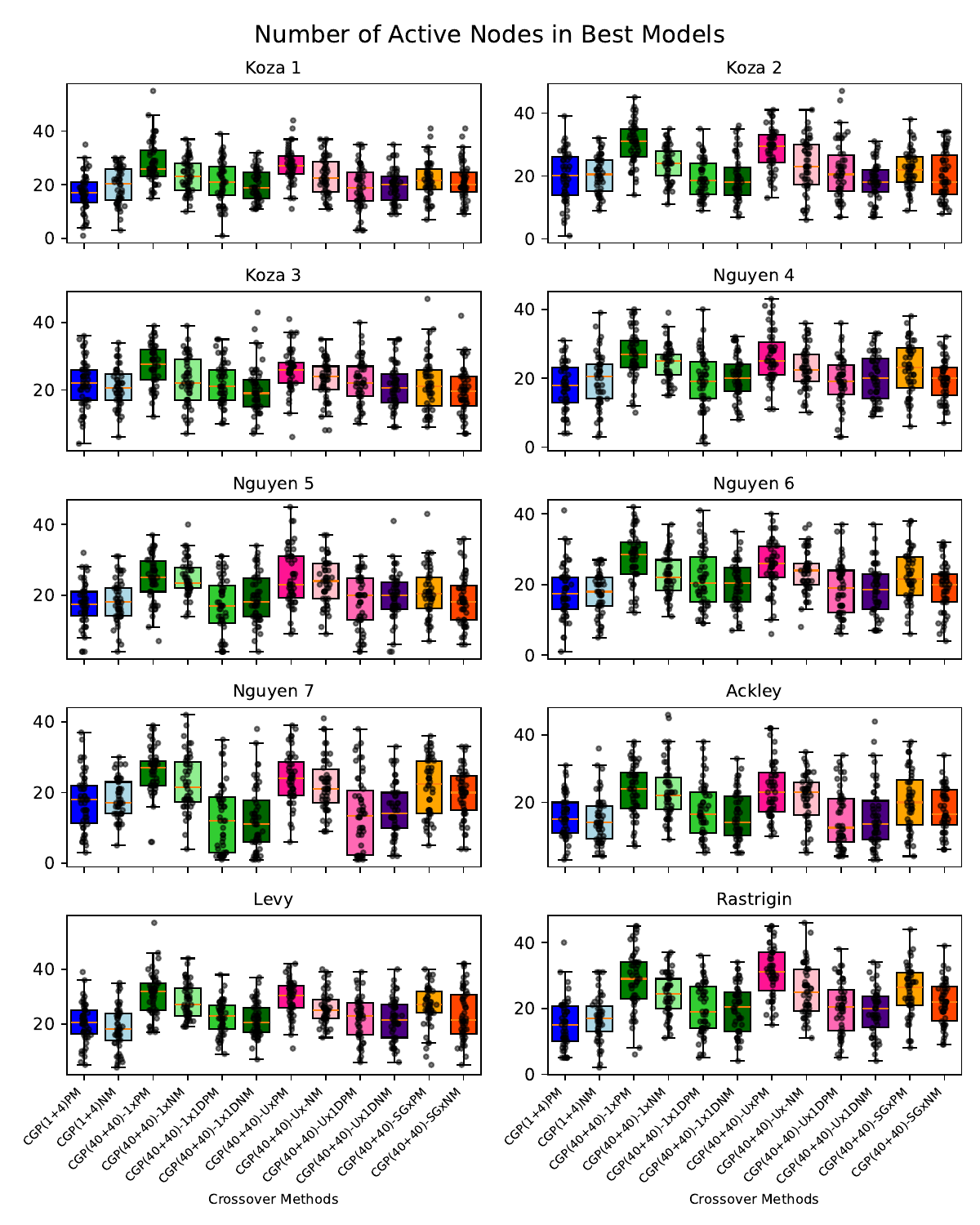}
    \caption{Box plots of the amount of active nodes in the best models at the end of evolution for each run.}
    \label{fig:sizes_compare}
\end{figure}

\section{Discussion}\label{sec:discussion}
In this work, we have demonstrated that forcing nodes to remain intact throughout the crossover process following~\cite{kalkreuth2020comprehensive} is a beneficial approach  for search on both synthetic and real world 
problems. We also propose that practitioners perform ``full'' node mutations by entirely replacing nodes rather than altering single genes, which we find to be beneficial for search in most cases.

What does this mean for how we see the history -- and future -- of CGP crossover research? It tells us that \textit{from the start}, the field has been subject to misconceptions. 
First, as Kocherovsky et al. demonstrated, the assumption that genes close to one another are functionally related is not applicable to CGP~\cite{kocherovsky2025on}. Second, the original design of CGP chromosomes as one-dimensional strings, an obvious inheritance from EC as a whole~\cite{miller1997designing,miller1999empirical} will evolve over generations, but treating each part of the instructions as their own genes 
does not result in productive operators. 

The subgraph crossover avoids this problem because its design \textit{requires} nodes to remain intact during crossover. Kalkreuth correctly assumed that an edge of the graph is not separable from the node it is directed at, and designed the operator to take this into account. Rewiring the first active node after the crossover point further serves to preserve good structures by partially repairing the destruction caused by one-point crossover.

Future research into CGP crossover should explicitly consider how their genomes are structured and why the researchers chose to structure them in that way. When new operators are developed, one should be cognizant of the fact that
\begin{itemize}
    \item There is a positional bias in node activity; it is more likely that nodes closer to the front will be active.

    \item Crossover is a destructive operation and it may be worth taking the time to repair part of the damage.
    
    \item Instructions close to each other in the genome are not necessarily -- and in fact are not likely to be -- related functionally. The very last node (node $n$) is as likely to take the first node as input as it is its immediate predecessor (node $n-1$).
\end{itemize}


Unless counter-measures are created for such tendencies, operators will suffer from the negative effects such tendencies can bring to an algorithm. We do not believe that the final word has been spoken on the efficiency of CGP operators, but think that node preservation holds a key role in designing them for productive application.

\clearpage

\bibliographystyle{splncs04}
\bibliography{references}

\end{document}